\documentclass[runningheads,a4paper]{llncs}

\usepackage{amssymb}
\usepackage{amsmath}
\usepackage{graphicx}
\usepackage{color}
\usepackage{booktabs}
\usepackage{subfigure}
\usepackage{url}
\usepackage{wrapfig}

\DeclareMathOperator{\Enc}{Enc}
\DeclareMathOperator{\Dec}{Dec}
\DeclareMathOperator{\Dis}{Dis}

\begin{document}

\mainmatter

\title{Deep Autoencoding Models for Unsupervised Anomaly Segmentation in Brain MR Images}
\titlerunning{Deep Unsupervised Anomaly Detection}

\author{Christoph Baur\inst{1} \and Benedikt Wiestler\inst{3} \and Shadi Albarqouni\inst{1} \and Nassir Navab\inst{1,2}}
\authorrunning{Baur et al.}
\institute{Computer Aided Medical Procedures (CAMP), TU Munich, Germany\\
\and Whiting School of Engineering, Johns Hopkins University, Baltimore, United States \and Neuroradiology Department, Klinikum Rechts der Isar, TU Munich, Germany}
\maketitle

\begin{abstract}
	Reliably modeling normality and differentiating abnormal appearances from normal cases is a very appealing approach for detecting pathologies in medical images. A plethora of such unsupervised anomaly detection approaches has been made in the medical domain, based on statistical methods, content-based retrieval, clustering and recently also deep learning. Previous approaches towards deep unsupervised anomaly detection model patches of normal anatomy with variants of Autoencoders or GANs, and detect anomalies either as outliers in the learned feature space or from large reconstruction errors. In contrast to these patch-based approaches, we show that deep spatial autoencoding models can be efficiently used to capture normal anatomical variability of entire 2D brain MR images. A variety of experiments on real MR data containing MS lesions corroborates our hypothesis that we can detect and even delineate anomalies in brain MR images by simply comparing input images to their reconstruction. Results show that constraints on the latent space and adversarial training can further improve the segmentation performance over standard deep representation learning.
	
\end{abstract}

\section{Introduction}
\begin{figure}[t]
\centering
\includegraphics[width=1.0\textwidth]{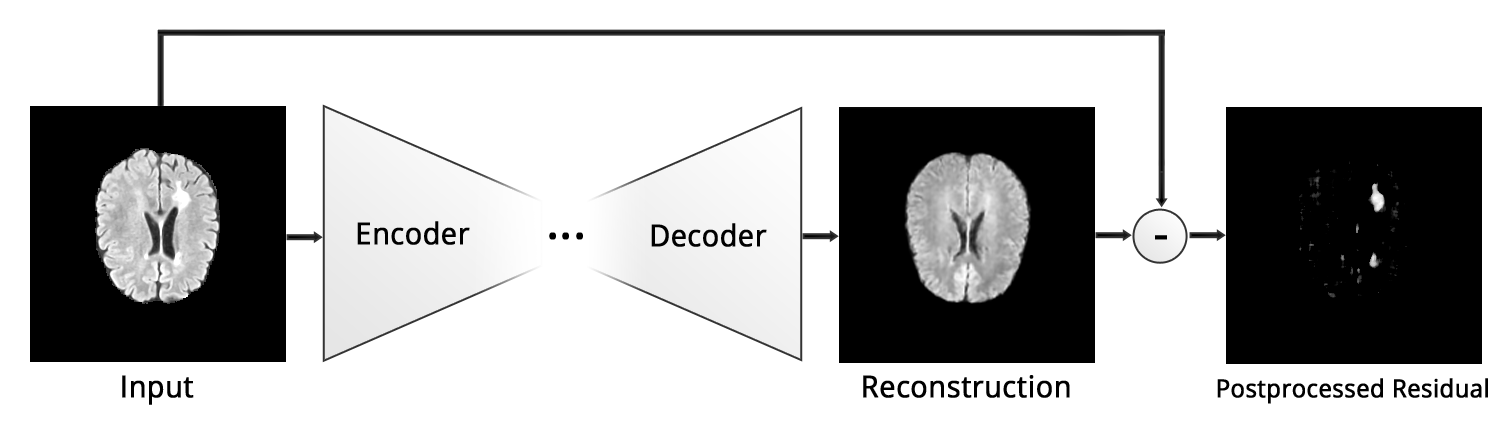}
\caption{The proposed anomaly detection oncept at a glance. A simple subtraction of the reconstructed image from the input reveals lesions in the brain.}
\label{fig:framework}
\end{figure}

Brain MR images are frequently acquired for detecting and diagnosing pathologies, monitoring disease progression and treatment planning. The manual identification and segmentation of pathologies in brain MR data is a tedious and time-consuming task. In an attempt to aid the detection and delineation of brain lesions arising from Multiple Sclerosis (MS), tumors or ischemias, the medical image analysis community has proposed a great variety of methods. Outstanding levels of performance have been achieved with recent supervised deep learning methods. However, their training requires vast amounts of labeled data which often is not available. Further, these approaches suffer from limited generalization since in general, training data rarely comprises the gamut of all possible pathological appearances~\cite{TaboadaCrispi:te}. Given the constrained anatomical variability of the healthy brain, an alternative approach is to model the distribution of healthy brains, and both detect and delineate pathologies as deviations from the norm. Here, we formulate the problem of brain lesion detection and delineation as an unsupervised anomaly detection (UAD) task based on state-of-the-art deep representation learning and adversarial training, requiring only a set of normal data and no labels at all. The detection and delineation of pathologies are thereby obtained from a pixel-wise reconstruction error (Fig. \ref{fig:framework}). To the best of our knowledge, this is the first application of deep convolutional representation learning for UAD in brain MR images which operates on entire MR slices.

\textbf{Related work} In the medical field, many efforts have been made towards UAD, which can be grouped into methods based on statistical modeling, content-based retrieval or clustering and outlier detection~\cite{TaboadaCrispi:te}. Weiss et al.~\cite{Weiss:2013cb} employed Dictionary Learning and Sparse Coding to learn a representation of normal brain patches in order to detect MS lesions. Other unsupervised MS lesion segmentation methods rely on thresholding and 3D connected component analysis~\cite{Iheme:2013hj} or fuzzy c-means clustering with topology constraints~\cite{Shiee:2010dw}. Notably, only few approaches have been made towards deep learning based UAD. Vaidhya et al.\cite{Vaidhya:2015im} utilized unsupervised 3D Stacked Denoising Autoencoders for patch-based glioma detection and segmentation in brain MR images, however only as a pretraining step for a supervised model. Recently, Schlegl et al.\cite{Schlegl:2017uq} presented the AnoGAN framework, in which they create a rich generative model of normal retinal OCT patches using a GAN. Assuming that the model cannot properly reconstruct abnormal samples, they classify query patches as either anomalous or normal by trying to optimize the latent code based on a novel mapping score, effectively also leading to a delineation of the anomalous region in the input data. In earlier work, Seeb{\"o}ck et al.\cite{Seebock:2016ug} trained an Autoencoder and utilized a one-class SVM on the compressed latent space to distinguish between normal and anomalous OCT patches. A plethora of work in the field of deep learning based UAD has been devoted to videos primarily based on Autoencoders (AEs) due to their ability to express non-linear transformations and the ability to detect anomalies directly from poor reconstructions of input data\cite{Sabokrou:2016gf,Hasan:2016gb,Chong:2017vb}. Very recently, first attempts have also been made with deep generative models such as Variational Autoencoders\cite{Kingma:2013tz,An:Ubdi4r8V} (VAEs), however limited to dense neural networks and 1D data. Noteworthy, most of this work focused on the detection rather than the delineation of anomalies.

A major advantage of AEs is their ability to reconstruct images with fairly high resolution thanks to a supervised training signal coming from the reconstruction objective. Unfortunately, they suffer from memorization and tend to produce blurry images. GANs\cite{Goodfellow:2014td} have shown to produce very sharp images due to adversarial training, however the training is very unstable and the generative process is prone to collapse to a few single samples. The recent formulation of VAEs has also shown that AEs can be turned into generative models which can mimic data distributions, and both concepts have also been combined into the VAEGAN\cite{Donahue:2016woa}, yielding a framework with the best of both worlds.

\textbf{Contribution} Inarguably, AnoGAN is a great concept for UAD in patch-based and small resolution scenarios, but as our experiments show, GANs lack the capability to reliably synthesize complex, high resolution brain MR images. Further, the approach requires a time-consuming iterative optimization. To overcome these issues, we propose AnoVAEGAN: We leverage a deep generative model in the form of spatial VAEs to build a model that captures ``global'' normal anatomical appearance rather than the variety of local patches. The reconstruction objective allows to train a generative model on complex, high resolution data such as brain MR slices. In order to avoid the memorization pitfalls of AEs and to improve realism of the reconstructed samples, we train the decoder part of the network with the help of an adversarial network, ultimately turning the model into a VAEGAN\cite{Donahue:2016woa}. In our experiments, we rank the AnoVAEGAN against the AnoGAN framework as well as both dense and spatial variants of the VAE, AE and Context Encoders\cite{Pathak:2016gb} (here referred to as ``AE-GAN'') in the tasks of unsupervised MS lesion delineation and report significant improvements of spatial autoencoding models over traditional ones.




\section{Methodology}
\label{sec:methodology}

\begin{figure}[t]
\centering
\includegraphics[width=1.0\textwidth]{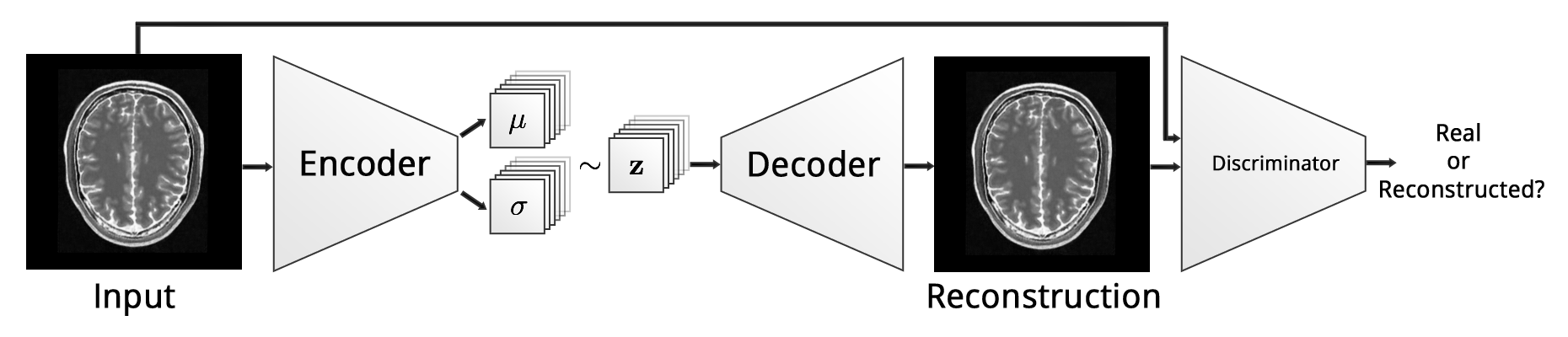}
\caption{An overview of our AnoVAEGAN}
\label{fig:sVAEGAN}
\end{figure}


In this work, we employ deep generative representation learning to model the distribution of the healthy brain, which should enable the model to fully reconstruct healthy brain anatomy while failing to reconstruct anomalous lesions in images of a diseased brain. Therefore, we utilize an adaptation of the VAEGAN\cite{Donahue:2016woa} to establish a parametric mapping from input images $\mathbf{x} \in \mathbb{R}^{H \times W}$ to a lower dimensional representation $\mathbf{z} \in \mathbb{R}^{d}$ and back to high quality image reconstructions $\mathbf{\hat{x}} \in \mathbb{R}^{H \times W}$ using an encoder $\Enc(\cdot; \theta)$ and a decoder $\Dec(\cdot; \phi)$:


\begin{equation}
	\mathbf{z} \sim \Enc(\mathbf{x}; \theta),\hspace{10pt} \mathbf{\hat{x}} = \Dec(\mathbf{z}; \phi), \hspace{10pt} \text{s.t.} \hspace{10pt} \mathbf{z} \sim \mathcal{N}(0,I)
\end{equation}

Like in \cite{Donahue:2016woa}, the latent space $\mathbf{z}$ is constrained to follow a multivariate normal distribution (MVN) $\mathcal{N}(0,I)$, which we leverage for encoding images of normal brain anatomy. Further, we employ a discriminator network $\Dis(\cdot)$ which classifies its input as either real or reconstructed. 

\subsubsection{Training.}
\label{ssub:training}
We optimize the framework using two loss functions in an alternating fashion. The VAE component of the model is optimized using

\begin{align*}
	\label{eq:vaeloss}
	\mathcal{L}_{VAE} &= \lambda_1 \mathcal{L}_{rec} + \lambda_2 \mathcal{L}_{prior} + \lambda_3 \mathcal{L}_{adv} \\
	&= \lambda_1 \|\mathbf{x}-\mathbf{\hat{x}}\|_1 + \lambda_2 \mathcal{D}_{KL}(\mathbf{z} || \mathcal{N}(0,I)) - \lambda_3 \log(\Dis(\Dec(\Enc(\mathbf{x})))),
\end{align*}
%


%
%

, while the discriminator is trained as commonly seen in the GAN framework\cite{Goodfellow:2014td}:

\begin{equation}
	\label{eq:disloss}
	\mathcal{L}_{Dis} = -\log(\Dis(\mathbf{x})) - \log(1-\Dis(\Dec(\Enc(\mathbf{z})))),
\end{equation}

Originally, VAEGAN used an abstract reconstruction loss in the latent space of the discriminator rather than a pixelwise reconstruction objective, which was not helpful for our purpose. For $\mathcal{L}_{rec}$, we thus used the pixelwise $\ell_1$-distance between input image and reconstruction. $\mathcal{L}_{prior}$ is the KL-Divergence between the distribution of generated $\mathbf{z}$ and a MVN, which is only used to regularize the weights $\theta$ of the encoder. The third part $\mathcal{L}_{adv}$ is the adversarial loss which forces the decoder to generate images that are likely to fool the discriminator in its task to distinguish between real and reconstructed images.

\begin{figure}[t]
\centering     
\subfigure[Dense Autoencoder dAE]{\label{fig:dAE}\includegraphics[width=60mm]{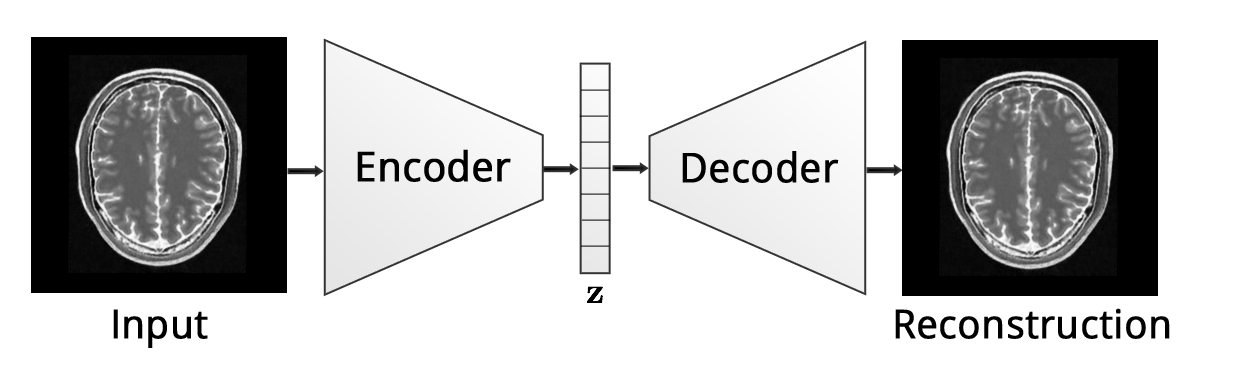}}
\subfigure[Spatial Autoencoder sAE]{\label{fig:sAE}\includegraphics[width=60mm]{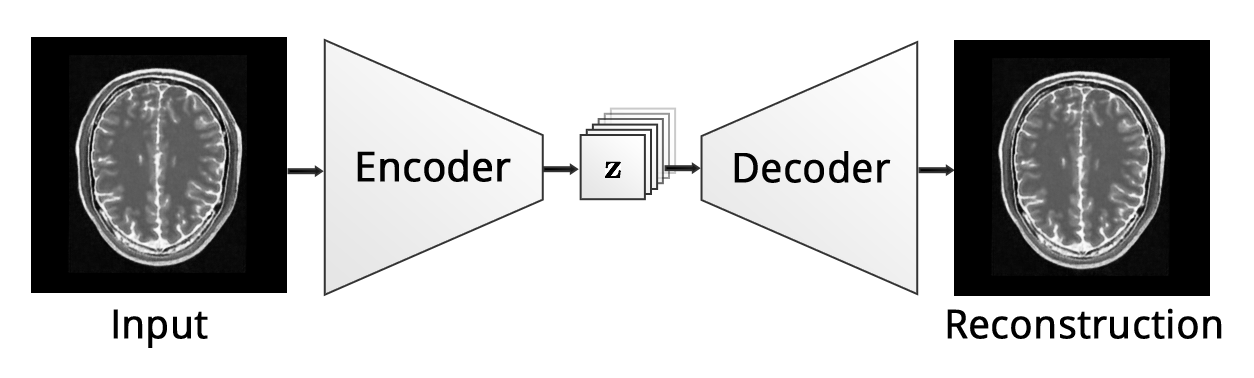}}
\subfigure[Dense Variational Autoencoder dVAE]{\label{fig:dVAE}\includegraphics[width=60mm]{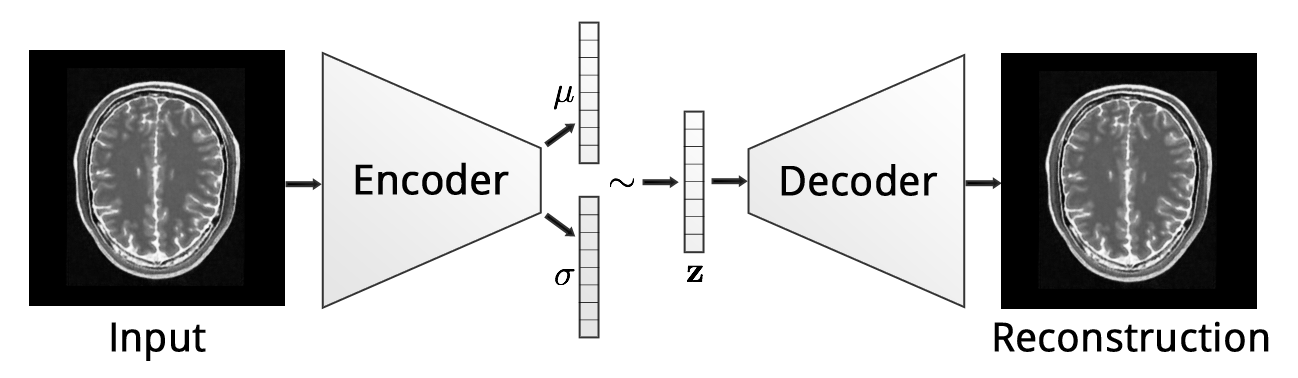}}
\subfigure[Spatial Variational Autoencoder sVAE]{\label{fig:sVAE}\includegraphics[width=60mm]{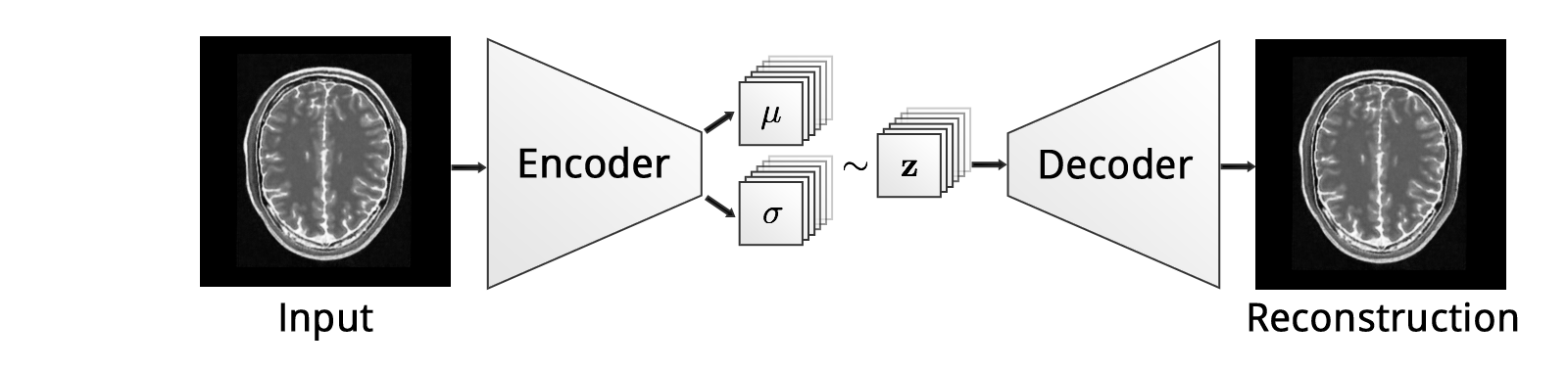}}
\caption{An overview of different Autoencoder frameworks}
\label{fig:aes}
\end{figure}

A peculiarity of our approach is the fully convolutional encoder-decoder architecture which we use in order to preserve spatial information in the latent space, i.e. $\mathbf{z} \in \mathbb{R}^{h\times w \times c}$ is a multidimensional tensor. Fig.~\ref{fig:sVAEGAN} shows our AnoVAEGAN, and a depiction of different AE architectures is given in Fig. \ref{fig:aes}.

\subsubsection{Anomaly Detection}
\label{ssub:anomaly_detection}
Anomalies are detected and delineated by 1) computing the pixelwise $\ell_1$-distance between an input image and its reconstruction and 2) thresholding the resulting residual image to obtain a binary segmentation.




\section{Experiments and Results}
\label{sec:real_experiments}
Given the variants of AE and our proposed framework, we investigate i) whether autoencoding deep networks can be utilized in general to learn to reconstruct complex brain MR images, ii) how the dimensionality of $\mathbf{z}$ affects the reconstruction capabilities of a model, iii) the effect of constraining $\mathbf{z}$ to be well structured and iv) if adversarial training enhances the quality of reconstructed images. In the following paragraphs we first introduce the dataset, provide implementational details and then describe the conducted experiments.

\paragraph{Datasets.} For our experiments, we use an inhouse dataset which provides a rich variety of images of healthy brain anatomy - a neccessity for our approach. The dataset consists of FLAIR and T1 image pairs from 83 patients (1360 slices) with healthy brains and 49 patients with MS lesions (980 slices). All images have been acquired with a Philips Achieva 3T scanner. To reduce variability and relax the reconstruction problem, all images have been rigidly co-registered to the SRI24 ATLAS~\cite{Rohlfing:2009dp}. Further, the skull has been stripped with ROBEX~\cite{Iglesias:2011fb}. The resulting images have been denoised using CurvatureFlow~\cite{Technology:tk} and normalized into the range [0,1]. From every patient, we extracted 20 consecutive axial slices of resolution $256 \times 256$px around the midline.

\begin{wrapfigure}[9]{l}{0.3\textwidth}
	\vspace{-25pt}
\centering
\includegraphics[width=0.3\textwidth]{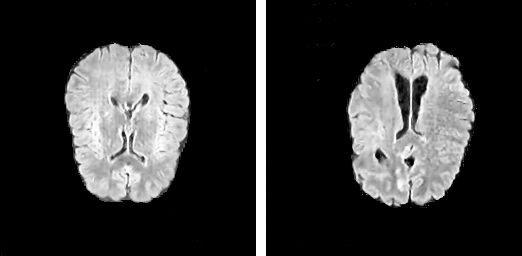}
\caption{Realistic (left) and unrealistic (right) samples generated with AnoGAN.}
\label{fig:anogan}
\end{wrapfigure}

\paragraph{Implementation.} We build upon the basic architecture proposed in~\cite{Donahue:2016woa} and perform only minor modifications affecting the latent space (see Table \ref{table:realResults}). Across different architectures we keep the model complexity of the encoder-decoder part the same to allow for a valid comparison. All models have been trained for 150 epochs in minibatches of size 8, using a learning rate of 0.001 for the reconstruction objective and 0.0001 for the adversarial training on a single nVidia 1080Ti GPU with 8GB of memory.

\paragraph{Evaluation Metrics.} We measure the performance of the different models by the mean and standard deviation of the Dice-Score across different testing patients as well as the average time required for reconstructing and segmenting a sample.

\begin{figure}[t]
\centering
\includegraphics[width=1.0\textwidth]{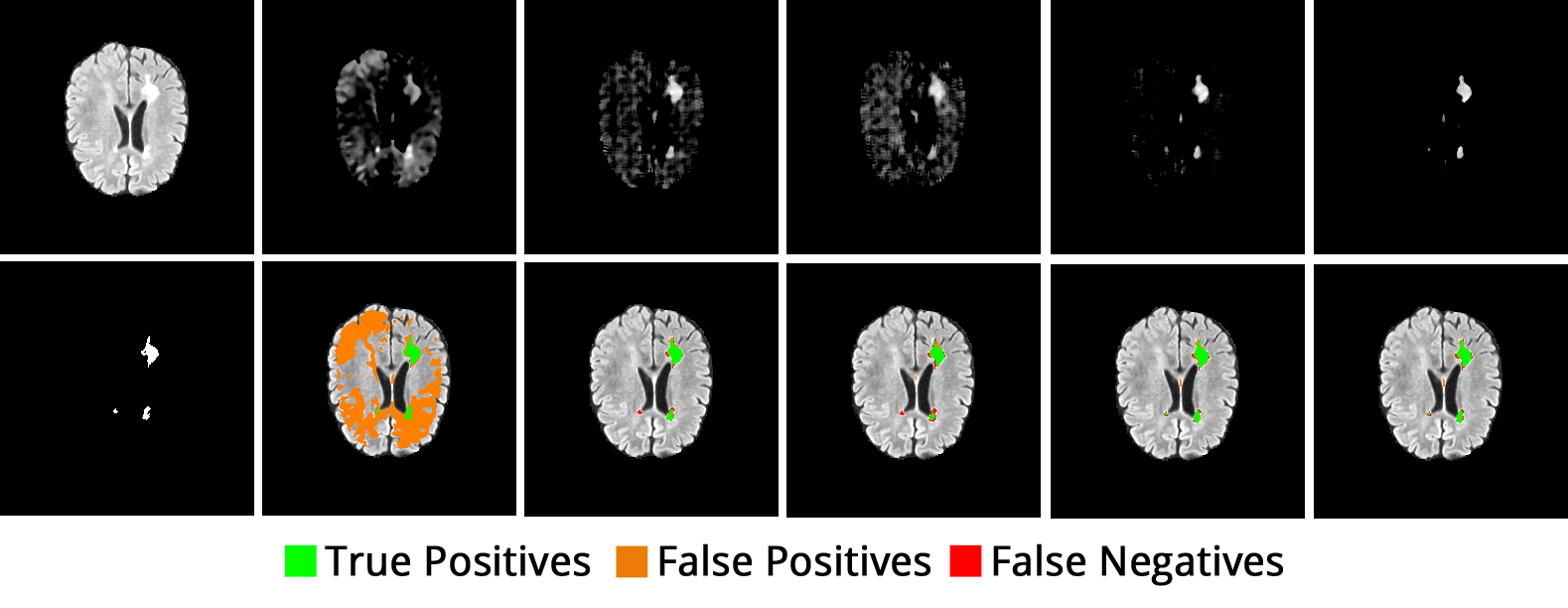}
\caption{1st Column: a selected axial slice and its ground-truth segmentation; Succeeding columns show the filtered difference images (top row) and the resulting segmentation augmented to the input image (bottom row) for the following models (in order): dAE, sAE$_3$, sAE-GAN, sVAE and sVAE-GAN.}
\label{fig:recs}
\end{figure}

\subsection{Anomaly Detection}
\label{sub:mskri_experiments}

We first trained normal convolutional AE \& VAE with a dense latent space of dimensionality $512$ and found that, besides not being capable of reconstructing brain lesions, they also lack the capability to reconstruct fine details such as the brain convolutions (Fig. \ref{fig:recs}). Similar to~\cite{Hasan:2016gb,Chong:2017vb}, we then make the architecture fully convolutional to ensure that spatial information is not lost in the bottleneck of the model. Notably, this heavily increases the dimensionality of $\mathbf{z}$. We thus vary the number of featuremaps of the spatial AE to investigate the impact on reconstruction quality of normal and anomalous samples. We identify $\mathbf{z}=16\times16\times64$ as a good parameterization and use it in further experiments on a spatial VAE, a spatial AE-GAN\cite{Pathak:2016gb} and our AnoVAEGAN. Further, we also trained an AnoGAN on the same set of normal axial slices for 150 epochs. After approx. 82 epochs of training, we obtained realistically looking images, however continuation of the training led to instabilities which resulted in unrealistic samples (Fig. \ref{fig:anogan}). Thus, we evaluated the AnoGAN approach after 82 epochs. The required iterative reconstruction of testing samples was computed in 100 steps.

\subsubsection{Postprocessing}
\label{ssub:postprocessing}

\begin{wrapfigure}[12]{l}{0.3\textwidth}
\vspace{-25pt}
\includegraphics[width=0.3\textwidth]{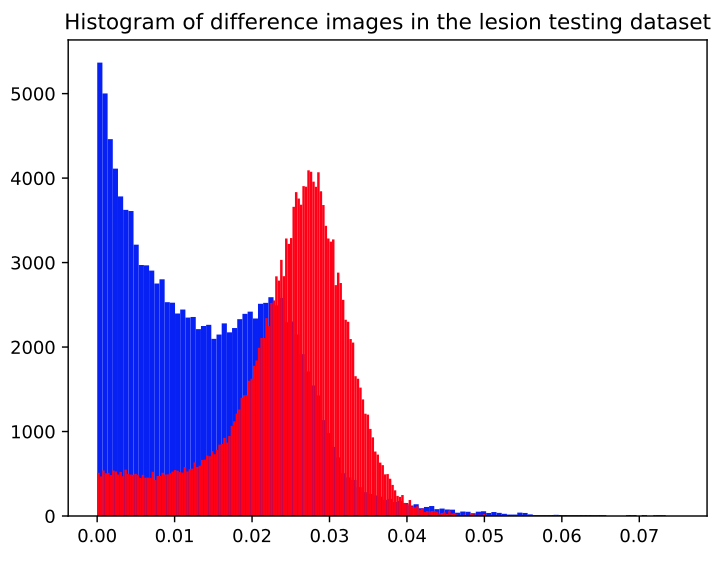}
\vspace{-25pt}
\caption{The histogram of residuals for normal (blue) and anomalous (red) pixels using our AnoVAEGAN.}
\label{fig:hist}
\end{wrapfigure}

After reconstruction of all the slices, we apply some postprocessing steps to reduce the number of False Positives. For every patient, we apply a $5\times5\times5$ median filter to the reconstructed subvolumes to filter out small residuals, usually belonging to brain convolutions. Further, we multiply the residuals with slightly eroded brain masks, threshold the resulting volumes to obtain a binary segmentation mask and remove tiny 3D connected components with an area less than 6 voxels as they are unlikely to constitute lesions. The threshold is model specific and determined as the 98th percentile of the models reconstruction errors on the training dataset. We chose this percentile empirically from the histogram of residuals obtained from both normal and abnormal data (Fig. \ref{fig:hist}). The performance of each model is reported in Table \ref{table:realResults}. A comparison of processed residual images and final segmentations of various models can be seen in Fig. \ref{fig:recs}.

The highest average Dice-score for MS lesion segmentation has been obtained with our AnoVAEGAN. The spatial VAEs and AEs which do not leverage adversarial training produce only slightly inferior scores, however. All spatial autoencoding models significantly outperform the ones with a dense bottleneck and, except for sAE$_1$, also the AnoGAN. Interestingly, the spatial AE with adversarial training performs worse than its generative counterpart and the other spatial AEs with the same bottleneck resolution.

\begin{table}[t]
\centering
\caption{Results of our experiments on unsupervised MS lesion segmentation. We report the Dice-Score (mean and std. deviation across patients) as well as the avg. reconstruction time per sample in seconds. Prefixes \emph{d} or \emph{s} stand for dense or spatial.}
\label{table:realResults}
\begin{tabular}{|l|l|l|l|}
\hline
MODELTYPE		& $\mathbf{z}$         	& DICE ($\mu \pm \sigma$)    					& avg. Reco.-time [s]	\\ \hline
dAE       	& $512$       					& $0.12764 \pm 0.14617$ 							& 0.01286	\\ \hline
sAE$_1$     & $8\times8\times64$    & $0.19739 \pm 0.19061$ 							&	0.01217 \\ \hline
sAE$_3$     & $16\times16\times64$  & $0.58558 \pm 0.19845$ 							&	0.01185 \\ \hline
sAE-GAN\cite{Pathak:2016gb} & $16\times16\times64$ & $0.52636 \pm 0.19780$ &	0.01445 \\ \hline
dVAE      	& $512$       					& $0.16619 \pm 0.17790$ 							&	0.01083 \\ \hline
sVAE      	& $16\times16\times64$  & $0.59227 \pm 0.19585$ 							&	0.01297 \\ \hline
AnoVAEGAN   & $16\times16\times64$  & $\mathbf{0.60508 \pm 0.19277}$ 			&	0.01541 \\ \hline
AnoGAN\cite{Schlegl:2017uq}  & $64$ & $0.37489 \pm 0.21926$ 							&	19.8547 \\ \hline

\end{tabular}
\end{table}



\section{Discussion and Conclusion}
\label{sec:conclusion}

Our experiments show that AE \& VAE models with dense bottlenecks cannot reconstruct anomalies, but at the same time lack the capability to reconstruct important fine details in brain MR images such as brain convolutions. By utilizing spatial AEs with sufficient bottleneck resolution, i.e. $16\times 16$px sized featuremaps, we can mitigate this problem. Noteworthy, a smaller bottleneck resolution of $8\times 8$px seems to lead to a severe information loss and thus to large reconstruction errors in general. By further constraining the latent space to follow a MVN distribution and introducing adversarial training, we notice marginal improvements over the non-generative models. As expected, spatial autoencoding clearly outperforms the AnoGAN and is considerably faster. While AnoGAN requires an iterative optimization, which consumes $\sim$19 seconds for a single reconstruction, all of the AE models require only a fraction of a second. Interestingly, even though the models operate on 2D data, the segmentations seem very consistent among neighboring axial slices.


In summary, we presented a novel and fast UAD approach based on generative deep representation learning which encodes the full context of brain MR slices. We believe that the approach does not only open up opportunities for unsupervised brain lesion segmentation, but can also act as prior information for supervised deep learning.

In future work, we also aim to investigate the projection of healthy anatomy into a latent space which follows a Gaussian Mixture Model rather than a single multivariate normal distribution, and intend to utilize 3D autoencoding models for unsupervised anomaly detection.

\section*{Acknowledgements}
\label{sec:acknowledgements}

We thank our clinical partners from Klinikum Rechts der Isar for providing us with their broad dataset of patients with healthy anatomy as well as the MS lesion dataset.

\bibliography{literature}

\begin{thebibliography}{10}
\providecommand{\url}[1]{\texttt{#1}}
\providecommand{\urlprefix}{URL }

\bibitem{An:Ubdi4r8V}
An, J., Cho, S.: {Variational Autoencoder based Anomaly Detection using
  Reconstruction Probability}. Tech. rep. (2015)

\bibitem{Chong:2017vb}
Chong, Y.S., Tay, Y.H.: {Abnormal Event Detection in Videos using
  Spatiotemporal Autoencoder.} CoRR  (2017)

\bibitem{Donahue:2016woa}
Donahue, J., Kr{\"a}henb{\"u}hl, P., Darrell, T.: {Adversarial Feature
  Learning}. arXiv.org  (May 2016)

\bibitem{Goodfellow:2014td}
Goodfellow, I.J., Pouget-Abadie, J., Mirza, M., Xu, B., Warde-Farley, D.,
  Ozair, S., Courville, A.C., Bengio, Y.: {Generative Adversarial Nets.} NIPS
  (2014)

\bibitem{Hasan:2016gb}
Hasan, M., Choi, J., Neumann, J., Roy-Chowdhury, A.K., Davis, L.S.: {Learning
  Temporal Regularity in Video Sequences}. In: 2016 IEEE Conference on Computer
  Vision and Pattern Recognition (CVPR). pp. 733--742. IEEE (2016)

\bibitem{Iglesias:2011fb}
Iglesias, J.E., Liu, C.Y., Thompson, P.M., Tu, Z.: {Robust Brain Extraction
  Across Datasets and Comparison With Publicly Available Methods}. IEEE
  Transactions on Medical Imaging  30(9),  1617--1634 (2011)

\bibitem{Iheme:2013hj}
Iheme, L.O., {\"U}nay, D., Baskaya, O., Sennaz, A., Kandemir, M., Yalciner,
  Z.B., Tepe, M.S., Kahraman, T., {\"U}nal, G.B.: {Concordance between
  computer-based neuroimaging findings and expert assessments in dementia
  grading.} SIU pp. 1--4 (2013)

\bibitem{Kingma:2013tz}
Kingma, D.P., Welling, M.: {Auto-Encoding Variational Bayes.} CoRR  (2013)

\bibitem{Pathak:2016gb}
Pathak, D., Kr{\"a}henb{\"u}hl, P., Donahue, J., Darrell, T., Efros, A.A.:
  {Context Encoders: Feature Learning by Inpainting}. In: 2016 IEEE Conference
  on Computer Vision and Pattern Recognition (CVPR). pp. 2536--2544. IEEE
  (2016)

\bibitem{Rohlfing:2009dp}
Rohlfing, T., Zahr, N.M., Sullivan, E.V., Pfefferbaum, A.: {The SRI24
  multichannel atlas of normal adult human brain structure}. Human Brain
  Mapping  31(5),  798--819 (Dec 2009)

\bibitem{Sabokrou:2016gf}
Sabokrou, M., Fathy, M., Hoseini, M.: {Video anomaly detection and localisation
  based on the sparsity and reconstruction error of auto-encoder}. Electronics
  Letters  52(13),  1122--1124 (Jun 2016)

\bibitem{Schlegl:2017uq}
Schlegl, T., Seeb{\"o}ck, P., Waldstein, S.M., Schmidt-Erfurth, U., Langs, G.:
  {Unsupervised Anomaly Detection with Generative Adversarial Networks to Guide
  Marker Discovery.} CoRR  cs.CV (2017)

\bibitem{Seebock:2016ug}
Seeb{\"o}ck, P., Waldstein, S.M., Klimscha, S., Gerendas, B.S., Donner, R.,
  Schlegl, T., Schmidt-Erfurth, U., Langs, G.: {Identifying and Categorizing
  Anomalies in Retinal Imaging Data.} CoRR  cs.LG (2016)

\bibitem{Technology:tk}
Sethian, J.A., et~al.: Level set methods and fast marching methods. Journal of
  Computing and Information Technology  11(1),  1--2 (2003)

\bibitem{Shiee:2010dw}
Shiee, N., Bazin, P.L., Ozturk, A., Reich, D.S., Calabresi, P.A., Pham, D.L.:
  {A topology-preserving approach to the segmentation of brain images with
  multiple sclerosis lesions.} NeuroImage  49(2),  1524--1535 (2010)

\bibitem{TaboadaCrispi:te}
Taboada-Crispi, A., Sahli, H.: {Anomaly Detection in Medical Image Analysis}.
  In: Handbook of Research on Advanced Techniques in Diagnostic Imaging and
  Biomedical Applications

\bibitem{Vaidhya:2015im}
Vaidhya, K., Thirunavukkarasu, S., Varghese, A., Krishnamurthi, G.:
  {Multi-modal Brain Tumor Segmentation Using Stacked Denoising Autoencoders.}
  Brainles@MICCAI  9556(10),  181--194 (2015)

\bibitem{Weiss:2013cb}
Weiss, N., Rueckert, D., Rao, A.: {Multiple Sclerosis Lesion Segmentation Using
  Dictionary Learning and Sparse Coding.} MICCAI  8149(Chapter 92),  735--742
  (2013)

\end{thebibliography}
\bibliographystyle{splncs03}

\end{document}